\documentclass[smallcondensed]{svjour3}

\usepackage{amsmath}
\usepackage{amssymb}

\newcommand{\e}[1]{\emph{#1}}

\newcommand{\mb}[1]{\mathbb{#1}}

\newcommand{\lra}{\longrightarrow}

\DeclareMathOperator{\NN}{NN}

\newcommand{\owa}[1]{\operatorname*{\mathnormal{#1\!\downarrow}}}

\DeclareMathOperator{\lp}{lp}
\DeclareMathOperator{\alp}{alp}

\usepackage[font=footnotesize]{subfig}
\usepackage{etoolbox}
\robustify\subref

\usepackage{booktabs}
\usepackage{longtable}
\usepackage{siunitx}
\usepackage{url}

\usepackage{tikz}
\usetikzlibrary{matrix,shapes,arrows,positioning,chains}

\usepackage[round]{natbib}
\setcitestyle{semicolumn,aysep={}}

\def\makeheadbox{\relax}

\usepackage[hidelinks]{hyperref}

\usepackage{fancyhdr}

\fancypagestyle{firstpagestyle}
{
\fancyhead[R]{\thepage\ifodd\value{page}\else\hfill\fi}
\fancyhf{}
\fancyhead[C]{\normalfont\scriptsize To appear in \href{https://doi.org/10.1007/s10994-022-06147-2}{Machine Learning 111.8 (2022) 2863--2883}. Submitted version.}
\fancyfoot[C]{\scriptsize © 2021. This manuscript version is made available under the \href{http://creativecommons.org/licenses/by-nc-nd/4.0/}{CC-BY-NC-ND 4.0 license}.}%
}

\begin{document}

\title{Optimised one-class classification performance}

\author{Oliver Urs Lenz \and Daniel Peralta \and Chris~Cornelis}
\date{}

\institute{O.U. Lenz \and C. Cornelis
\at Department of Applied Mathematics, Computer Science and Statistics, Ghent University, Belgium\\
\email{\{oliver.lenz,chris.cornelis\}@ugent.be}
\and
D. Peralta
\at IDLab, Department of Information Technology, Ghent University -- imec, Belgium\\
\email{daniel.peralta@ugent.be}
}

\maketitle

\thispagestyle{firstpagestyle}

\begin{abstract}
We provide a thorough treatment of one-class classification with hyperparameter optimisation for five data descriptors: Support Vector Machine (SVM), Nearest Neighbour Distance (NND), Localised Nearest Neighbour Distance (LNND), Local Outlier Factor (LOF) and Average Localised Proximity (ALP). The hyperparameters of SVM and LOF have to be optimised through cross-validation, while NND, LNND and ALP allow an efficient form of leave-one-out validation and the reuse of a single nearest-neighbour query. We experimentally evaluate the effect of hyperparameter optimisation with 246 classification problems drawn from 50 datasets. From a selection of optimisation algorithms, the recent Malherbe-Powell proposal optimises the hyperparameters of all data descriptors most efficiently. We calculate the increase in test AUROC and the amount of overfitting as a function of the number of hyperparameter evaluations. After 50 evaluations, ALP and SVM significantly outperform LOF, NND and LNND, and LOF and NND outperform LNND. The performance of ALP and SVM is comparable, but ALP can be optimised more efficiently so constitutes a good default choice. Alternatively, using validation AUROC as a selection criterion between ALP or SVM gives the best overall result, and NND is the least computationally demanding option. We thus end up with a clear trade-off between three choices, allowing practitioners to make an informed decision.

\keywords{
Data descriptors \and
Hyperparameter optimisation \and
Novelty detection \and
One-class classification \and
Semi-supervised outlier detection
}
\end{abstract}

\section{Introduction}
The goal of one-class classification (\citet{tax01oneclass}, also known as \e{novelty}, \e{semi-supervised outlier} or \e{semi-supervised anomaly detection}), is to form, on the basis of a representative sample, a model of a target (or \e{positive}) class that can later be used to predict whether unseen instances belong to that target class. Such one-class classification algorithms are called data descriptors. The difference with binary classification and the closely related \e{learning from positive and unlabelled data} \citep{bekker20learning} lies in the fact that a data descriptor only uses training data belonging to the target class.

In recent years, one-class classification has been applied to a wide range of problems, including the detection of tweets promoting hate or extremism \citep{agarwal15using}, or generated by bots \citep{rodriguezruiz20oneclass}, user authentication based on keystroke dynamics \citep{antal15evaluation}, writer identification \citep{hadjadji18two}, detecting abnormal train door operations \citep{ribeiro16sequential}, and the identification of different tumor cell subtypes \citep{sokolov16oneclass}.

A number of popular data descriptors require setting one or more hyperparameters. In a recent paper \citep{lenz21average}, we have identified optimal default hyperparameter values for a selection of these, and evaluated their performance. These default values allow the data descriptors to be used on one-class classification problems without negative training samples. However, for problems where a (limited) sample of negative data is available, this data can be used to optimise, or `tune', the hyperparameter values. Notably, for all of the concrete applications of one-class classification listed in the previous paragraph, there was negative data that could have been used for hyperparameter optimisation. In addition, negative data is always available when data descriptors are used as building blocks in a multi-class classification ensemble \citep{ban06implementing}.

The aim of this paper is to provide empirical evidence for the potential performance benefits of hyperparameter optimisation. The performance of data descriptors with hyperparameter optimisation has previously been evaluated by \citet{janssens09outlier} and \citet{swersky16evaluation}. In the present article, we go further, and try to answer the question: ``What is the best way to optimise the hyperparameter values of a data descriptor?''. Whereas both previous works use a naive grid search to optimise hyperparameter values, we evaluate a representative selection of optimisation algorithms, and identify the most effective strategy. We also explain how data descriptors based on nearest neighbour searches can be optimised using efficient leave-one-out validation instead of ordinary cross-validation.

A second question that we address is: ``How long should hyperparameter optimisation run for?'' Besides being inefficient, grid search also requires fixing the total number of evaluations before optimisation begins. In contrast, the optimisation algorithms that we consider sequentially select points in the hyperparameter space to evaluate. This allows us to present our results in terms of the number of evaluations, giving practitioners more insight into the effect of this choice.

We compare the data descriptors from \citet{lenz21average} that have one or more optimisable hyperparameters: Average Local Proximity (ALP), Support Vector Machine (SVM), Nearest Neighbour Distance (NND), Localised Nearest Neighbour Distance (LNND) and Local Outlier Factor (LOF). 

In the experiments performed by \citet{janssens09outlier}, SVM and LOF were tied, ahead of LNND, but the difference was not statistically significant. \citet{swersky16evaluation} ranked SVM, NND, LOF and LNND from high to low in that order, but only the difference between SVM and LNND was statistically significant. The former study did not include NND, and neither study included ALP, which had not yet been proposed at the time. Thus, ALP remains untested in the context of hyperparameter optimisation, whereas we found in \citet{lenz21average} that it is the best-performing data descriptor with default hyperparameter values (although the difference with SVM was only weakly significant).

Both previous studies evaluated performance with a Nemenyi test on mean ranks \citep{demsar06statistical}. This forced them to amalgamate results from one-class classification problems derived from the same dataset, of which there were 24 in \citet{janssens09outlier} and 30 in \citet{swersky16evaluation}. In order to get more statistical certainty, we draw from a larger number of datasets (50), and we compare pairs of data descriptors using a clustered Wilcoxon signed rank test \citep{rosner06wilcoxon} that allows us to use the full results from all 246 one-class classification problems. This choice is additionally motivated by the criticism that the $p$-value generated by the Nemenyi test for a pair of machine learning algorithms is too strongly dependent on the inclusion of other algorithms in the comparison \citep{benavoli16should}.

Together, these improvements over the previous studies --- using a suitable optimisation procedure, more datasets and a more precise statistical test, and including ALP --- allow us to provide a stronger answer to our third and final question: ``What is the best data descriptor for one-class classification with hyperparameter optimisation?''

We proceed by discussing the various data descriptors (Section \ref{sec_data_descriptors}) and optimisation algorithms (Section \ref{sec_optimisation}) in this article, explaining how our experiments are structured (Section \ref{sec_setup}), discussing the results of these experiments (Section \ref{sec_results}) and presenting our conclusions (Section \ref{sec_conclusion}).

\section{Data descriptors}
\label{sec_data_descriptors}
In this section, we list the data descriptors under consideration in the present paper. For a more detailed discussion, the reader is referred to \citet{lenz21average}. Here we focus on how the hyperparameters of these data descriptors can be optimised. The goal in each case is to maximise the area under the receiver operator characteristic (AUROC).

\subsection{Support Vector Machine}
The Support Vector Machine (SVM) data descriptor was proposed independently by \citet{tax99data,tax99support,tax04support} and \citet{scholkopf99estimating,scholkopf01estimating} as an adaptation of the soft-margin Support Vector Machine for binary classification \citep{cortes95support}. Both variants fit a surface to isolate the training data in the feature space, and allow application of the kernel trick to transform the feature space. The recommended kernel is the Gaussian kernel, parametrised with a value $c \in (0, \infty)$. With the Gaussian kernel, the two variants become equivalent. We use the Schölkopf variant, which solves an optimisation problem to fit a hyperplane between most of the target data and the origin, at a maximal distance to the origin. A hyperparameter $\nu \in (0, 1]$ controls the relative weight placed respectively on maximising distance to the origin, and not leaving training instances on the same side as the origin.

Thus, there are two hyperparameters that have to be chosen, and can be optimised with training data: $\nu$ and $c$. Because many of the optimisation methods that we consider require compact hyperparameter domains, we reparametrise $c$ as $\frac{c'}{1 - c'}$, and optimise $c'$ in $[10^{-6}, 1 - 10^{-6}]$, and restrict the domain of $\nu$ to $[10^{-6}, 1]$.

In order to optimise $\nu$ and $c'$, we apply stratified five-fold cross-validation to obtain five splits of the available training data into a smaller training set and a validation set. For each training set, we select the target class instances to obtain a target set. We evaluate a pair of hyperparameter values by constructing a model on the target set, querying with the respective validation sets, and calculating the mean of the resulting AUROC scores. This means that optimising the hyperparameters of SVM requires constructing five models for each pair of values to be evaluated.

\subsection{Nearest Neighbour Distance}
Nearest Neighbour Distance (NND) is a much simpler data descriptor than SVM, and goes back to at least \citet{knorr97unified}; it classifies instances in accordance with the distance to their $k$th nearest neighbour in the target data. In principle, the distance measure can be chosen freely, but in order to allow the efficient form of optimisation discussed below, we fix this choice to the Manhattan metric, which generally gives better results than the Euclidean metric \citep{lenz21average}. This leaves $k$ as the only hyperparameter to be optimised. Since it encodes a magnitude, we optimise $k$ logarithmically. To avoid having to work with extremely large arrays, and knowing that its optimal default value is simply 1, we limit $k$ to $\min(n, 100 \log n)$.

Because NND is so simple, $k$ can be optimised more efficiently than the hyperparameters of SVM. Firstly, it is not necessary to completely recalculate a new NND model for each value of $k$. if $k_{\max}$ is the maximum value for $k$ that we want to consider, we only require a single sorted $k_{\max}$ nearest neighbour query, since this contains all $k$th nearest neighbours for $k \leq k_{\max}$.

Secondly, instead of five-fold cross-validation, we use an efficient form of leave-one-out cross-validation, where each validation set contains a single instance. For five-fold cross-validation, we would have to create five nearest neighbour search models, one for each target set corresponding to a fold. To perform leave-one-out cross-validation, we create a single nearest neighbour search model on the basis of all target set instances. For each target class instance, we must ensure that it is not also part of the target set corresponding to its fold, so we query to obtain its $k_{\max} + 1$ nearest neighbours, and remove the first nearest neighbour distance (with value 0). For other instances, there is nothing to correct and we can simply perform a $k_{\max}$ nearest neighbour query. We collect the resulting scores and calculate a single validation AUROC.

\subsection{Localised Nearest Neighbour Distance}
Localised Nearest Neighbour Distance (LNND) \citep{ridderde98experimental,tax98outlier} is a simple extension of NND. It divides the $k$th nearest neighbour distance of an instance by that $k$th nearest neighbour's own $k$th nearest neighbour distance. This is motivated by the observation that the typical distance between neighbours in the target set may vary throughout the feature space, and that the nearest neighbour distance of a test instance should be relativised accordingly.

As with NND, we adopt the Manhattan metric and optimise $k$ logarithmically, up to $\min(n, 100 \log n)$.

We can also perform efficient leave-one-out cross-validation as with NND, but we have to do some additional work to obtain a correct result. For each target class instance, we have to check whether it is among the $k$ nearest neighbours of its $k$th nearest target class neighbour, and if so, substitute the $k+1$th nearest neighbour.

\subsection{Local Outlier Factor}
Local Outlier Factor (LOF) \citep{breunig00lof} is a more complex realisation of the idea behind LNND: it relativises the so-called \e{local reachability density} of a test instance against the local reachability density of its $k$ nearest neighbours.

Again as with NND and LNND, we adopt the Manhattan metric and optimise $k$ logarithmically, up to $\min(n, 100 \log n)$.

The calculation of LOF requires determining the $k$th nearest neighbour distance of the $i$th nearest neighbour of the $j$th nearest neighbour of a test instance (for all $i, j \leq k$). For this reason, it is no longer feasible to apply the efficient form of leave-one-out cross-validation described for NND and LNND, and we resort to performing stratified five-fold cross-validation as with SVM. However, we still retain the efficiency that for each fold, we only need one query to obtain the $k_{\max}$ nearest neighbours of a test instance.

\subsection{Average Localised Proximity}

Average Localised Proximity (ALP) \citep{lenz21average} aims to be more robust than LNND but less complex than LOF.

For a choice of integers $k$ and $l$, the average localised proximity of an instance $y$ is defined as follows. For each $i \leq k$, and each $j \leq l$, we establish $d_i(\NN_j(y))$, the $i$th neighbour distance of the $j$th nearest neighbour of $y$. We then take, for each $i$, the weighted mean of these values, to obtain the local $i$th neighbour distance relative to $y$:
\begin{equation}
D_i(y) = \sum_{j \leq l}w^l_j \cdot d_i(\NN_j(y)),
\end{equation}
and use this to relativise the $i$th nearest neighbour distance of $y$ in the target data $d_i(y)$, resulting in the $i$th localised proximity of $y$:
\begin{equation}
\lp_i(y) = \frac{D_i(y)}{D_i(y) + d_i(y)}.
\end{equation}
Finally, we aggregate over the $k$ localised proximities of $y$ by applying the ordered weighted average $\owa{w^k}$ \citep{yager88ordered}, which is the weighted mean of the values sorted from large to small:
\begin{equation}
\alp(y) = \owa{w^k}_{i \leq k} \lp_i(y).
\end{equation}

The distance measure used is the Manhattan metric, and the weights $w^k$ and $w^l$ are linearly decreasing: $\frac{p}{p(p + 1)/2}, \frac{p - 1}{p(p + 1)/2}, \dots, \frac{1}{p(p + 1)/2}$, for $p = k, l$. This leaves $k$ and $l$ to be optimised, which respectively determine the scale at which nearest neighbour distances are considered and the amount of localisation. As with NND, we optimise $k$ and $l$ logarithmically.

Similar to NND and LNND, we only need a single $\max(k_{\max}, l_{\max})$ nearest neighbour query if we want to evaluate values of $k$ and $l$ up to $k_{\max}$ and $l_{\max}$. We can also apply efficient leave-one-out validation, by performing essentially the same correction as for LNND. In particular, we correct the local distances of a neighbour $x$ of $y$ by considering its $k + 1$ nearest neighbours, and removing either the distance to $y$ or the $k + 1$th nearest neighbour distance.

Increasing $k$ and $l$ has two effects: it draws in more distant neighbours of $y$, and it decreases the slope of the weight vectors, making the contribution of successive neighbours more equal. The asymptotic limit of this process is a weight vector with equal weights everywhere. A more practical limit is that $k$ and $l$ cannot grow beyond the number of target instances $n$. However, we can simulate higher values for $k$ and $l$ by truncating the weight vector and multiplying by a constant to ensure that its sum still equals 1.

This also allows us to address a computational issue with large datasets, that evaluating a pair of values for $k$ and $l$ on the whole training set involves $(k + 1) \cdot l \cdot n$ distance values. If all three values are large, processing a single array with all distance values requires a very large amount of memory. For these reasons, we let $k$ and $l$ range up to $5n$, but truncate distance values and weight vectors after $\min(n, 20 \log n)$. This is informed by the knowledge that the optimal default values for $k$ and $l$ are $5.5 \log n$ and $6 \log n$ respectively \citep{lenz21average}.

\section{Optimisation algorithms}
\label{sec_optimisation}
Optimisation problems are typically formulated in terms of a problem function $f: P \lra \mb{R}$, where the \e{problem space} $P$ is a subspace of $\mb{R}^m$ for some $m$ that is often required to be compact. Depending on the context, the goal of optimisation is to find points in $P$ that minimise or maximise $f$. In the present article, we wish to maximise the validation AUROC of our data descriptors, and the problem space is determined by the hyperparameters that we optimise. 

There is an important distinction between algorithms that aim to find a local optimum of $f$ in $P$, and those that aim to identify the global optimum of $f$ in $P$. An essential characteristic of global optimisation algorithms is that they have to balance the exploration of areas of the problem space with large uncertainty and the exploitation of areas where function performance is known to be good.

An intrinsic disadvantage of local optimisation algorithms is that they require a choice of starting point, and may get stuck in a local optimum if this starting point is chosen poorly. However, the optimisation problems of finding the best hyperparameter values for our data descriptors may be close to unimodal, in which case this risk could be relatively small. For this reason, we include two classical local optimisation algorithms that are easy to implement.

\subsection{Random search}
Purely random search is a surprisingly potent global optimisation strategy. By continuing to evaluate arbitrary points in the problem space, we can expect to eventually get arbitrarily close to the global optimum. In particular, random search is a more efficient algorithm than grid search \citep{bergstra12random}. Because this strategy uses no information from previous evaluations, it serves as a good baseline.

\subsection{Hooke-Jeeves}
The local search algorithm proposed by \citet{hooke61direct} passes through the problem space in steps. Each step follows a pattern, which is the vector sum of a number of substeps along each coordinate axis. The pattern is adjusted with each step, by optionally adding or subtracting a substep along each coordinate axis, depending on which option results in the greatest decrease in the objective function value. If the pattern cannot be adjusted to produce a step with any improvement, a new pattern is created from scratch. If no such pattern can be found either, the substep size is decreased.

We use the implementation provided by pymoo \citep{blank20pymoo} and its default values. As starting values we use the optimal default values identified in \citet{lenz21average}.

\subsection{Nelder-Mead}
The local search algorithm proposed by \citet{nelder65simplex} is based on an earlier proposal by \citet{spendley62sequential}. It iterates on $m + 1$ points that can be viewed as the vertices of an $m$-dimensional simplex. The algorithm lets this simplex `walk' through the problem space by replacing its worst vertex in each iteration. Each step is directed towards the mid-point of the remaining vertices. The new vertex is either placed between the worst vertex and this mid-point (shrinking the simplex), or beyond the midpoint (reflecting and optionally extending it). If none of these options improve upon the worst vertex, the entire simplex is shrunk, with only the best vertex remaining in place. 

The theoretical performance of Nelder-Mead optimisation had long been unclear, until \citet{torczon89multidirectional} demonstrated with a concrete example that Nelder-Mead can converge on points that are not local optima, even with problems that are twice differentiable. Nevertheless, Nelder-Mead optimisation has been very popular due to its relative simplicity, and because it seems to converge very quickly in many simple practical applications \citep{wright95direct}.

We use the implementation provided by SciPy \citep{virtanen20scipy}, with starting simplices centred around the optimal default values identified in \citet{lenz21average}.

\subsection{Kushner-Sittler}
A global optimisation method was first proposed by \citet{kushner62versatile,kushner64new}, based in part on unpublished work by Robert A Sittler (see \citet{betro91bayesian}, \citet{jones01taxonomy} and \citet{brochu09tutorial} for overviews of later developments). This has been referred to as simply \e{global optimisation}, or \e{Bayesian optimisation}, because its central idea is to iteratively use the Bayesian information criterion to select the next point in the problem space to evaluate. We assume that the unknown problem function is drawn from a random distribution of functions, which is traditionally modelled as a Gaussian process. In each step, we can calculate the conditional probability $p(y|x)$ that the problem function will obtain certain values at a given point $x$ in the problem space, in light of the function values that have already been calculated. These conditional probabilities are then reduced to a single score for each $x$ with an activation function, and the point with the maximal such score is selected as the next point to evaluate. The activation function most often used today, already hinted at in \citet{kushner62versatile}, is \e{expected improvement}, the expected value of $\max(y - y^*, 0)$ under the model for some $y^* \in \mb{R}$, typically the largest evaluated function value so far.

Kushner-Sittler optimisation transforms the original optimisation problem into a series of new optimisation problems for each iteration. These new optimisations incur a certain cost themselves, but this cost is only dependent on the dimensionality of the original problem, so for problems that are costly to evaluate, the trade-off is worthwhile. Note also that as with the original problem, these optima can generally only be approximated, but it is (often tacitly) assumed that this is not problematic.

We use the implementation provided by Emukit \citep{paleyes19emulation}, with the first point chosen randomly.

\subsection{Bergstra-Bardenet}
A more recent variant of Kushner-Sittler optimisation, motivated in particular by high-dimensional and conditional problem spaces, is the Tree-structured Parzen Estimator (TPE) proposal by \citet{bergstra11algorithms}. It lets the target value $y^*$ correspond to a quantile of the evaluated function values. This induces a split between small and large values, and the corresponding distributions $p(y < y^*)$ and $p(y > y^*)$, which can be modelled with two Parzen Estimators $l(x)$ and $g(x)$. By reformulating $p(y|x)$ in terms of $p(x|y)$, $p(y)$ and $p(x)$, the authors then show that the expected improvement of the original model is maximal when $\frac{g(x)}{l(x)}$ is maximal.

We use the Adaptive TPE implementation of hyperopt \citep{bergstra11algorithms}.

\subsection{Malherbe-Powell}
Global optimisation algorithms often proceed from the assumption that the problem function satisfies certain smoothness conditions. In particular, for any $k > 0$, we can define the class of $k$-Lipschitz functions as those functions $f$ that satisfy:

\begin{align}
\forall x_1, x_2 \in A: \left\lvert f(x_1) - f(x_2) \right\rvert \leq k \cdot \left\lvert x_1 - x_2 \right\rvert \label{eq_lipschitz}
\end{align}

\citet{malherbe17global} propose that we can use this assumption to restrict the search to certain parts of the problem space. They propose the LIPO algorithm, in which random candidates are drawn from the problem space, but only those candidates are evaluated that potentially improve upon the current optimum, in view of the candidates evaluated so far and (\ref{eq_lipschitz}).

In general, $k$ may be difficult to estimate, and we simply want to assume that some such $k$ exists for a problem function. For this purpose, \citet{malherbe17global} also propose the AdaLIPO algorithm. This alternates LIPO with purely random search, and increases $k$ whenever (\ref{eq_lipschitz}) is no longer satisfied.

AdaLIPO was further modified into MaxLIPO and implemented into the dlib library by \citet{king09dlibml, king17global}. MaxLIPO presents three improvements. Firstly, it incorporates a noise term that prevents $k$ from approaching infinity if the problem function is not in fact $k$-Lipschitz because it has small discontinuities. Secondly, it employs separate values of $k$ for each dimension. And thirdly, instead of selecting new candidates at random, it identifies the candidate with the largest potential improvement in light of (\ref{eq_lipschitz}).

A more fundamental problem of AdaLIPO that carries over to MaxLIPO is that while it is seemingly able to quickly locate the neighbourhood of the global optimum, it then takes much longer to approach the optimum itself. This can be understood, since by considering the maximal potential improvement, rather than some form of expected improvement, these algorithms place a greater emphasis on exploration than exploitation. To address this, \citet{king17global} lets MaxLIPO alternate with the trust region approach of the local optimiser BOBYQA \citep{powell09bobyqa}, a bounded version of the earlier NEWUOA proposal \citep{powell04newuoa}.

\section{Experimental setup}
\label{sec_setup}

In order to enable comparison with using default hyperparameter values, we closely follow the experimental setup of \citet{lenz21average}. We use the same collection of 246 one-class classification problems, drawn from 50 datasets from the UCI machine learning repository \citep{dua19uci}, rescaled through division by the interquartile range \citep{rousseeuw93alternatives} of each feature in the target class. We have implemented the data descriptors in our own open-source Python wrapper fuzzy-rough-learn\footnote{\url{https://github.com/oulenz/fuzzy-rough-learn}} \citep{lenz20fuzzyroughlearn}, which uses scikit-learn \citep{pedregosa11scikitlearn} as a backend for SVM and nearest neighbour queries.

For each one-class classification problem, we apply five-fold cross-validation. We measure the performance of a data descriptor with specific hyperparameter values in terms of the area under the receiver operator curve (AUROC). For each division, we measure validation AUROC using nested cross- or leave-one-out validation as explained in Section \ref{sec_data_descriptors}, as well as test AUROC by retraining the data descriptor on all of the training data and evaluating its performance on the test data.

We maximise validation AUROC by applying the optimisation algorithms from Section \ref{sec_optimisation}. Each optimiser is allowed a maximum budget of 50 evaluations of hyperparameter values. Although the NND, LNND, LOF and ALP hyperparameters $k$ and $l$ are discrete, in order to be able to apply the selected optimisation algorithms, we optimise them as if they were continuous.\footnote{We note that more generally, for any classification algorithm, performance metrics like AUROC, accuracy, precision and recall are non-continuous because a finite number of instances can only be subjected to a finite number of rankings.} As a result, subsequent steps of the optimisation search may target different points in the problem space that are discretised back to the same concrete value(s), which are not evaluated again. Thus, with local optimisers that have reached a local optimum, as well as with small datasets in general, the optimisation search may never reach 50 evaluations. Therefore, we terminate all optimisation searches after 100 steps.

We structure our analysis in terms of the number of evaluations. For each cross-validation division and number of evaluations, we use the hyperparameter values that maximise validation AUROC up to that point. We start our analysis by identifying the most suitable optimiser for each data descriptor, and adopt this for the rest of our analysis. We compare data descriptors both by summarising performance with the mean test AUROC, and by looking at individual results at the level of cross-validation divisions. In both cases, we apply a weighting scheme such that the  50 datasets from which the one-classification problems are drawn all contribute equally. In scatter plots, this is reflected in the size of the markers. We measure rank correlation with the weighted Kendall's $\tau$ \citep{vigna15weighted}.

In order to determine statistical significance, we apply clustered Wilcoxon signed-rank tests \citep{rosner06wilcoxon} on the mean AUROC scores across folds for each one-class classification problem. When we compare data descriptors to each other, we use the Holm-Bonferroni method \citep{holm79simple} to correct for family-wise error.

\section{Results and analysis}
\label{sec_results}

\begin{figure}
\centering
\includegraphics[width=\linewidth]{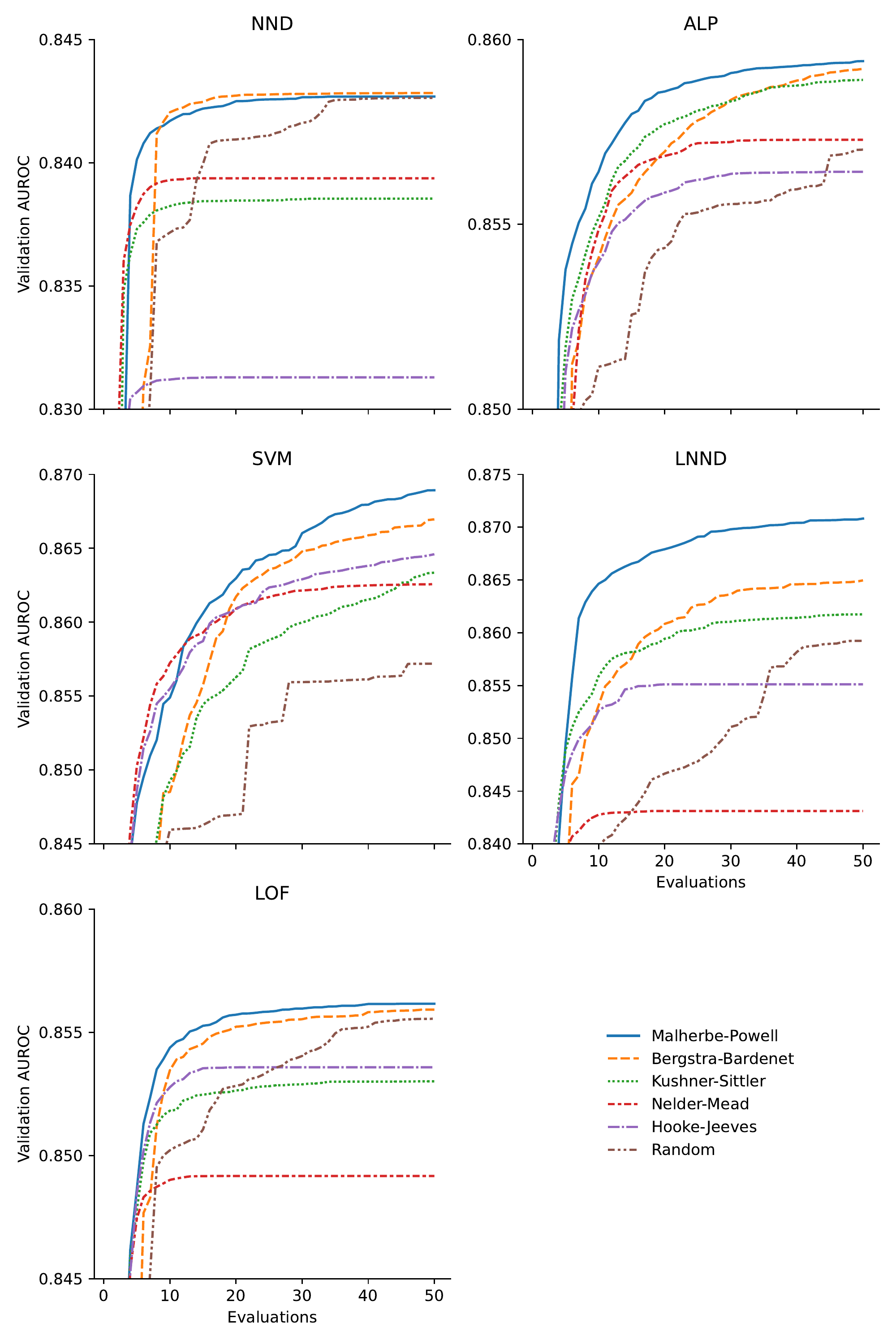}%
\caption{Weighted mean validation AUROC of data descriptors with hyperparameters tuned with a number of different optimisers.}
\label{fig_optimisation_validation_auroc}
\end{figure}

Figure \ref{fig_optimisation_validation_auroc} shows the performance of the different optimisers with the data descriptors. We can get an idea of the difficulty of these optimisation problems by looking at the baseline random strategy. For NND, our maximum budget of fifty evaluations is essentially enough for random search to find the global maxima, while for LOF, it comes reasonably close. LNND and ALP are more difficult, and random search clearly lags behind the other optimisation strategies with SVM. Taking into account the dimensionality of the respective problem spaces, it appears that the problem curves of LNND and SVM are relatively hard to optimise.

It is clear that the two local optimisation methods, Nelder-Mead and Hooke-Jeeves, generally fail to find the global optima because they get stuck in a local optimum. Neither method performs clearly better than the other. Nevertheless, if ease of implementation is a larger priority than performance, they may be an acceptable option for ALP and SVM. For the data descriptors with one hyperparameter, simple random search is to be preferred.

Of the global algorithms, the performance of Kushner-Sittler is surprisingly poor, in particular with NND and LOF, where it appears to stall below the level reached by the best local algorithm. Closer inspection reveals that it is too strongly focused on exploitation over exploration, and will often evaluate long series of points in the problem space that are very close together. This may be due to the chosen implementation, or the fact that the hyperparameters $k$ and $l$ are locally constant.

The overall best-performing method is Malherbe-Powell, with Bergstra-Bardenet in clear second place. Malherbe-Powell finds the highest AUROC for all data descriptors, except NND, where the difference with Bergstra-Bardenet is minimal ($1.4 \cdot 10^{-4}$). Therefore, we will use the results of Malherbe-Powell for the rest of this section. However, we also note that some of the differences are very small, and practitioners may want to prioritise ease of implementation when selecting an optimiser.

\begin{figure}
\centering
\includegraphics[width=\linewidth]{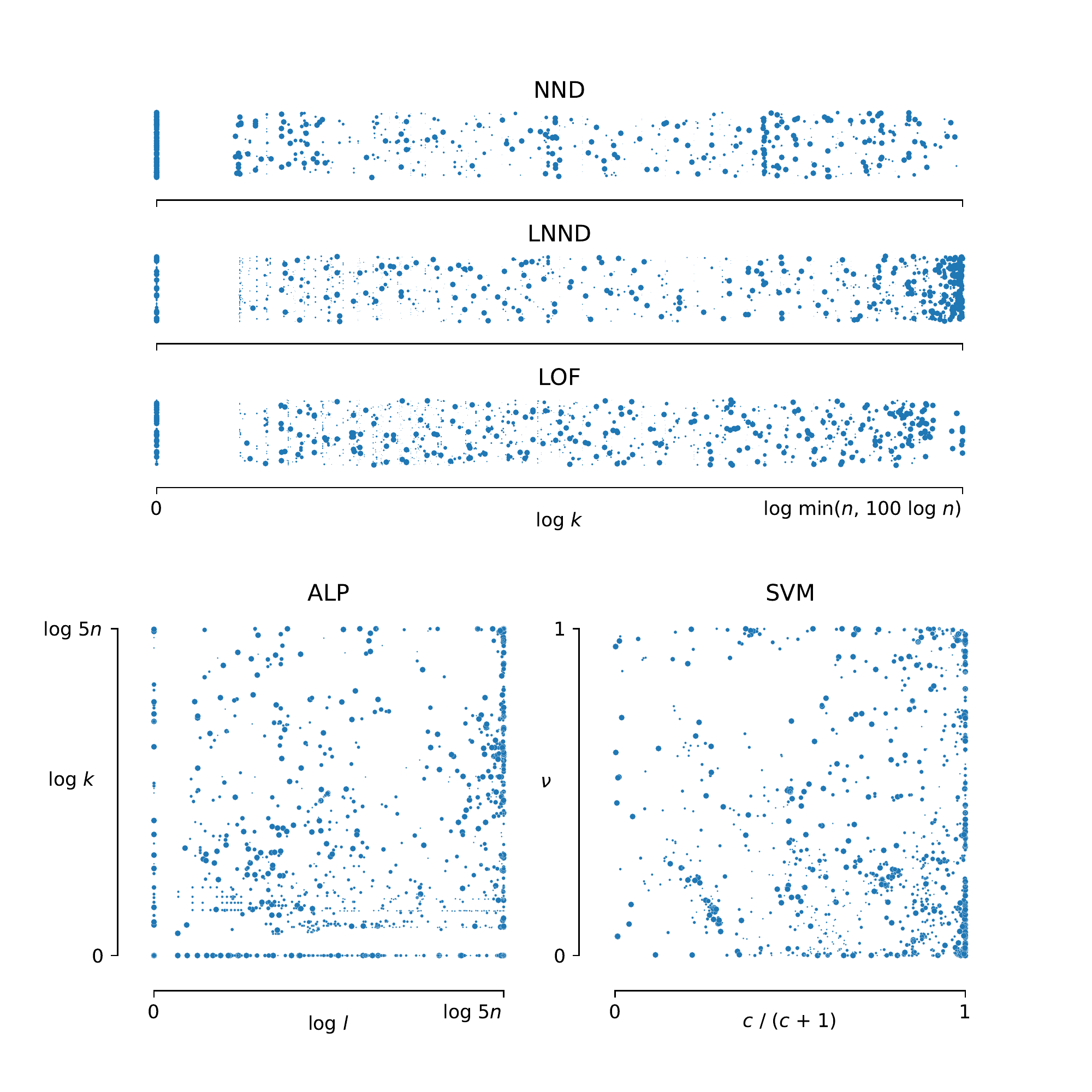}%
\caption{Distribution of optimised hyperparameter values after 50 evaluations.}
\label{fig_optimised_param_vals}
\end{figure}

Figure \ref{fig_optimised_param_vals} shows the distribution of the hyperparameter values after 50 evaluations. These distributions are relatively uniform, which suggests that the chosen parametrisations are efficient. For LOF (7.3\%), LNND (8.6\%) and especially NND (39\%), there is a substantial minority of problems for which the optimal value is simply 1. In the case of NND, for which 1 is the default value, this indicates that $k$ often doesn't need to be optimised. For LNND, a large number of optimal values (23\%) are within 1\% of the maximum, but this is in most cases due to the fact that $k$ cannot increase beyond $n$, rather than our imposed limit of $100 \log n$.

\begin{figure}
\centering
\includegraphics[width=\linewidth]{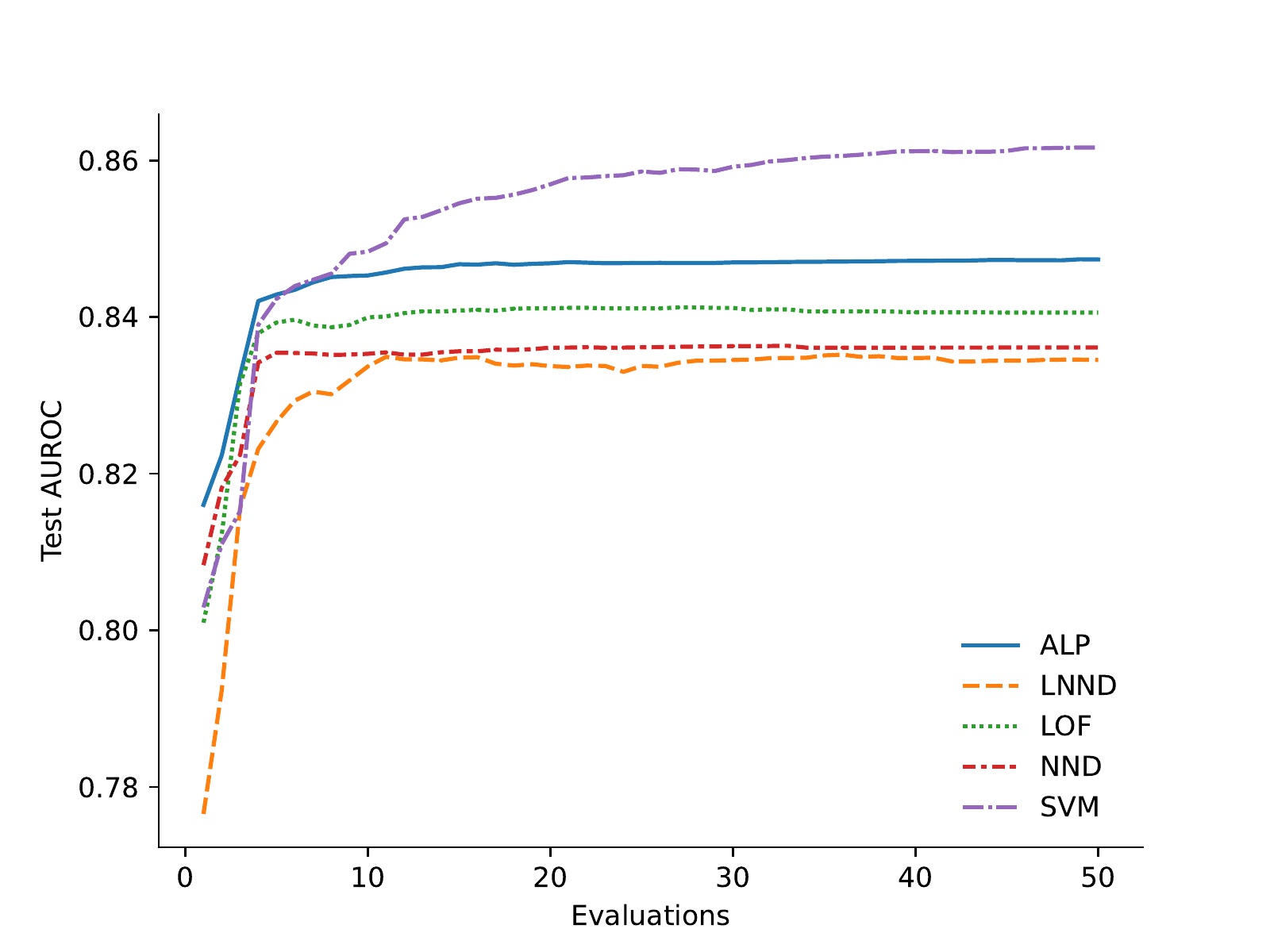}%
\caption{Weighted mean test AUROC.}
\label{fig_test_auroc}
\end{figure}

\begin{figure}
\centering
\includegraphics[width=\linewidth]{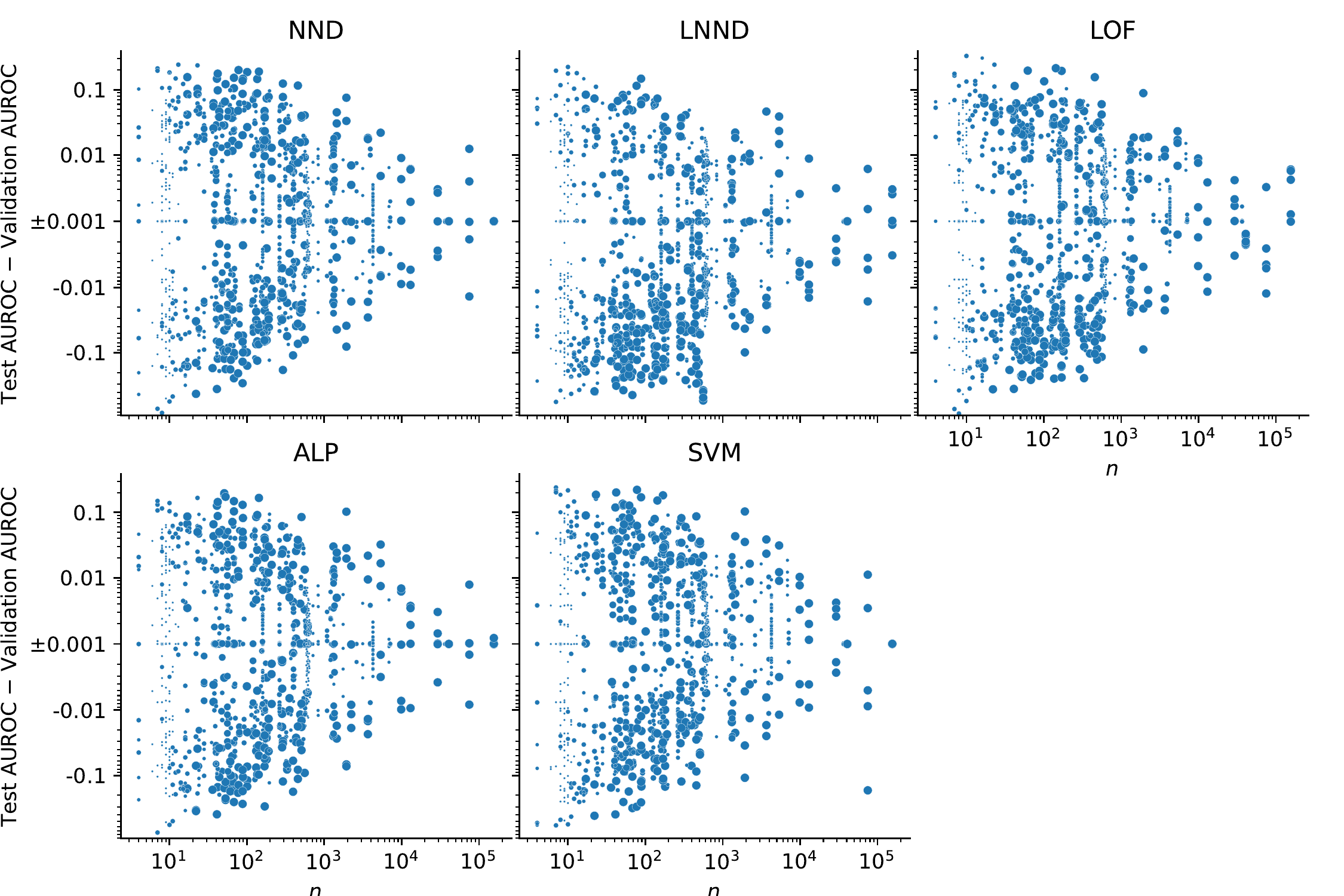}%
\caption{Difference between test and validation AUROC as a function of target class size $n$.}
\label{fig_overfitting_size}
\end{figure}

Figure \ref{fig_test_auroc} shows the weighted mean test AUROC of the data descriptors. The data descriptors display varying sensitivity to hyperparameter optimisation. All test AUROC curves increase steeply for 4 evaluations and then flatten out. However, for SVM, the initial rise is steeper and its curve continues to increase for much longer, allowing it to surpass LOF, NND and ALP even though it starts quite low. LNND improves even more steeply, but because it starts out very poorly, remains the worst-performing data descriptor. The data descriptors approach their final scores (after 50 evaluations) to within 0.001 points after respectively 5 (NND), 10 (LNND and LOF), 13 (ALP) and 37 (SVM) evaluations. 

The test AUROC curves in Figure \ref{fig_test_auroc} don't achieve the same scores as the validation AUROC curves for Malherbe-Powell optimisation in Figure \ref{fig_optimisation_validation_auroc}. The difference, which can be interpreted as overfitting, is largest for LNND (0.036 after fifty evaluations), followed by LOF (0.016), ALP (0.012), SVM (0.0073) and NND (0.0066). Note that LOF and LNND, with one hyperparameter each, show more overfitting than SVM and ALP, with two hyperparameters. The weighted Kendall's $\tau$ for the amount of overfitting after 50 evaluations ranges from 0.31 for SVM and LNND to 0.59 for SVM and NND. The amount of overfitting should be taken into account when estimating test AUROC or selecting data descriptors on the basis of validation AUROC. However, the numbers cited above are dependent on the mix of datasets that we use: for all data descriptors, validation AUROC becomes increasingly accurate as the target set size grows (Figure \ref{fig_overfitting_size}).

\begin{figure}
\centering
\includegraphics[width=\linewidth]{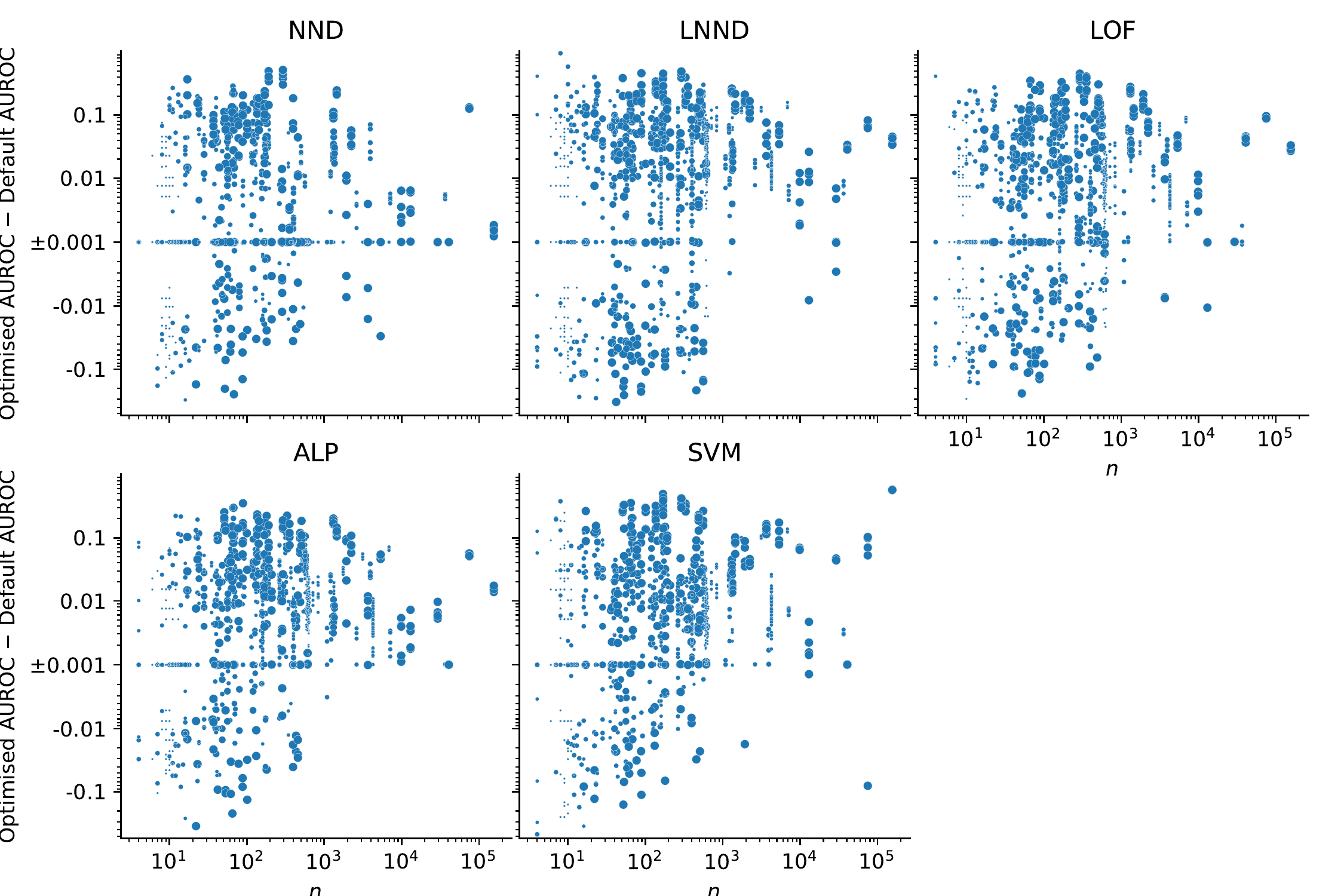}%
\caption{Increased or decreased AUROC due to hyperparameter optimisation over default hyperparameter values, as a function of target class size $n$.}
\label{fig_tuned_default_diff_by_size}
\end{figure}

When we apply a clustered Wilcoxon signed rank test to compare the test AUROC obtained with hyperparameter optimisation and the test AUROC obtained with default hyperparameter values from \citet{lenz21average}, we find that optimised values start to outperform default values with great certainty ($p < 0.01$) within 2 (LNND, ALP), 3 (LOF, SVM) and 4 (NND) evaluations. With optimised values, SVM, LOF and NND also perform significantly better ($p < 0.01$) than ALP, the best data descriptor with default values, after 4 (SVM), 5 (LOF) and 20 (NND) evaluations. Even after fifty evaluations, LNND with optimised values still performs worse than ALP with default values. Hyperparameter optimisation is not guaranteed to increase AUROC for any of the data descriptors, especially with smaller datasets (Figure \ref{fig_tuned_default_diff_by_size}).

\begin{figure}
\centering
\includegraphics[width=\linewidth]{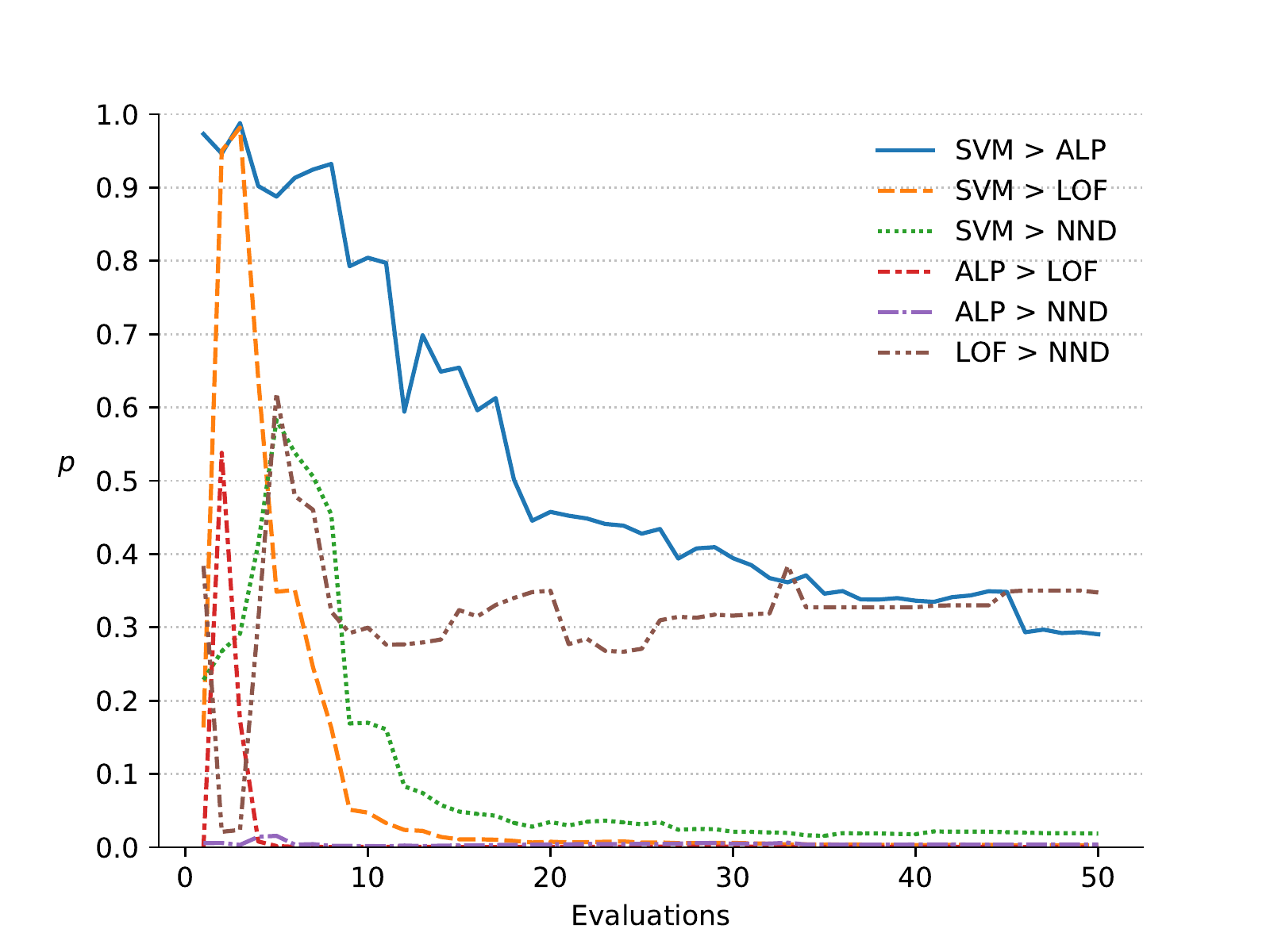}%
\caption{One-sided $p$-values of clustered Wilcoxon signed rank tests that one data descriptor is better than another (uncorrected for multiple testing).}
\label{fig_p_values}
\end{figure}

\begin{table}
\centering
\caption{One-sided $p$-values of clustered Wilcoxon signed-rank tests of AUROC after 50 evaluations, testing row data descriptor $>$ column data descriptor, with Holm-Bonferroni family-wise error correction applied to each row.}
\label{tab_p_values}
\begin{tabular}{llllll}
\toprule
    &    ALP &       LOF &      NND &       LNND \\
\midrule
SVM & $0.29$ &  $0.0079$ &  $0.037$ & $< 0.0001$ \\
ALP &        & $0.00091$ & $0.0077$ & $< 0.0001$ \\
LOF &        &           & $\geq 1$ & $< 0.0001$ \\
NND &        &           &          &  $0.00037$ \\
\bottomrule
\end{tabular}
\end{table}

The test AUROC scores after 50 evaluations are highly rank-correlated: the weighted Kendall's $\tau$ ranges from 0.74 (SVM and LNND) to 0.86 (SVM and NND). To determine whether the differences in performance are statistically significant, we perform one-sided clustered Wilcoxon signed rank tests. The resulting $p$-values after each evaluation are displayed in Figure \ref{fig_p_values}. The $p$-value for the opposite test can be obtained by subtracting the respective value from 1. Tests with LNND are omitted from Figure \ref{fig_p_values} since the corresponding $p$-values don't rise above 0.01. Table \ref{tab_p_values} lists the $p$-values after 50 evaluations, corrected for multiple testing.

Based on these experiments, we can confidently say that with sufficient evaluations, ALP and SVM perform better than NND, LOF and LNND, and that NND and LOF also perform better than LNND. We have far less certainty about the relative performance of ALP and SVM, and of NND and LOF. Figure \ref{fig_p_values} suggests that LOF generally outperforms NND, and that when the number of evaluations is small, ALP outperforms SVM, and vice-versa when the number of evaluations is large, but there is a large possibility that these observations are simply due to chance.

\begin{table}
\centering
\caption{Fraction of one-class classification problems with higher training and test AUROC by ALP or SVM.}
\label{tab_train_test_alp_svm_cm}
\begin{tabular}{lllll}
\toprule
Validation AUROC & Test AUROC\\
\cmidrule{2-5}
{} & ALP $<$ SVM & ALP $=$ SVM & ALP $>$ SVM &   Total \\
\midrule
ALP $<$ SVM &      $0.37$ &     $0.031$ &     $0.088$ &  $0.49$ \\
ALP $=$ SVM &   $0.00080$ &     $0.033$ &    $0.0052$ & $0.039$ \\
ALP $>$ SVM &      $0.12$ &     $0.038$ &      $0.31$ &  $0.47$ \\
Total       &      $0.49$ &      $0.10$ &      $0.40$ &         \\
\bottomrule
\end{tabular}
\end{table}

If we focus on the performance after 50 evaluations (Table \ref{tab_train_test_alp_svm_cm}), we see that SVM obtains a higher AUROC than ALP slightly more often than vice-versa, both on validation and test data. This confirms that the average performance of ALP and SVM is very close in practice. However, for individual classification problems, the choice still matters. We note that which data descriptor performs better is fairly consistent between validation and test data. If for each classification problem, we choose the data descriptor that obtains a higher validation AUROC (choosing ALP in event of a tie), it will perform worse on test data in only 21\% of cases. The advantage of this combination of ALP and SVM over either of ALP or SVM on its own is highly significant ($p < 0.0001$), regardless of whether we choose for each fold separately or on the basis of the mean validation AUROC across folds.

\begin{figure}
\centering
\includegraphics[width=\linewidth]{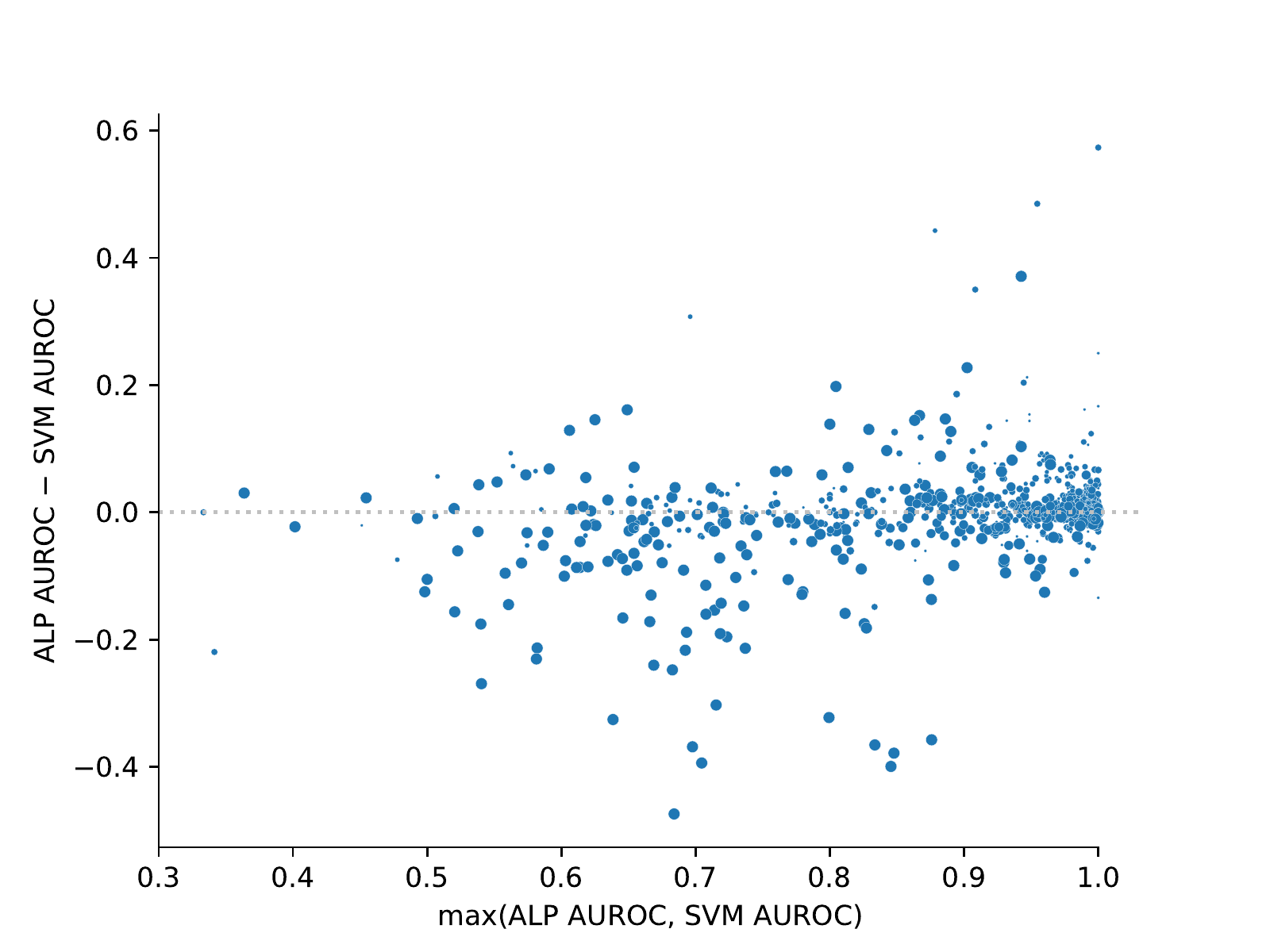}%
\caption{Difference between ALP and SVM AUROC, as a function of the difficulty of one-class classification problems, expressed by the maximum of ALP and SVM AUROC.}
\label{fig_diff_vs_max_auroc}
\end{figure}

One factor that plays a role in the relative performance of SVM and ALP is the difficulty of the one-class classification problem. For the purpose of the present analysis, we can express this as the maximum of the AUROC achieved by ALP and SVM. Figure \ref{fig_diff_vs_max_auroc} plots the relative performance of ALP and SVM against this difficulty. SVM is better able to separate more difficult problems, but for problems for which a good AUROC of 0.8 or more can be achieved, ALP beats SVM more often (46\%) than vice-versa (41\%), with a weighted mean difference of 0.0021.

\begin{figure}
\centering
\includegraphics[width=\linewidth]{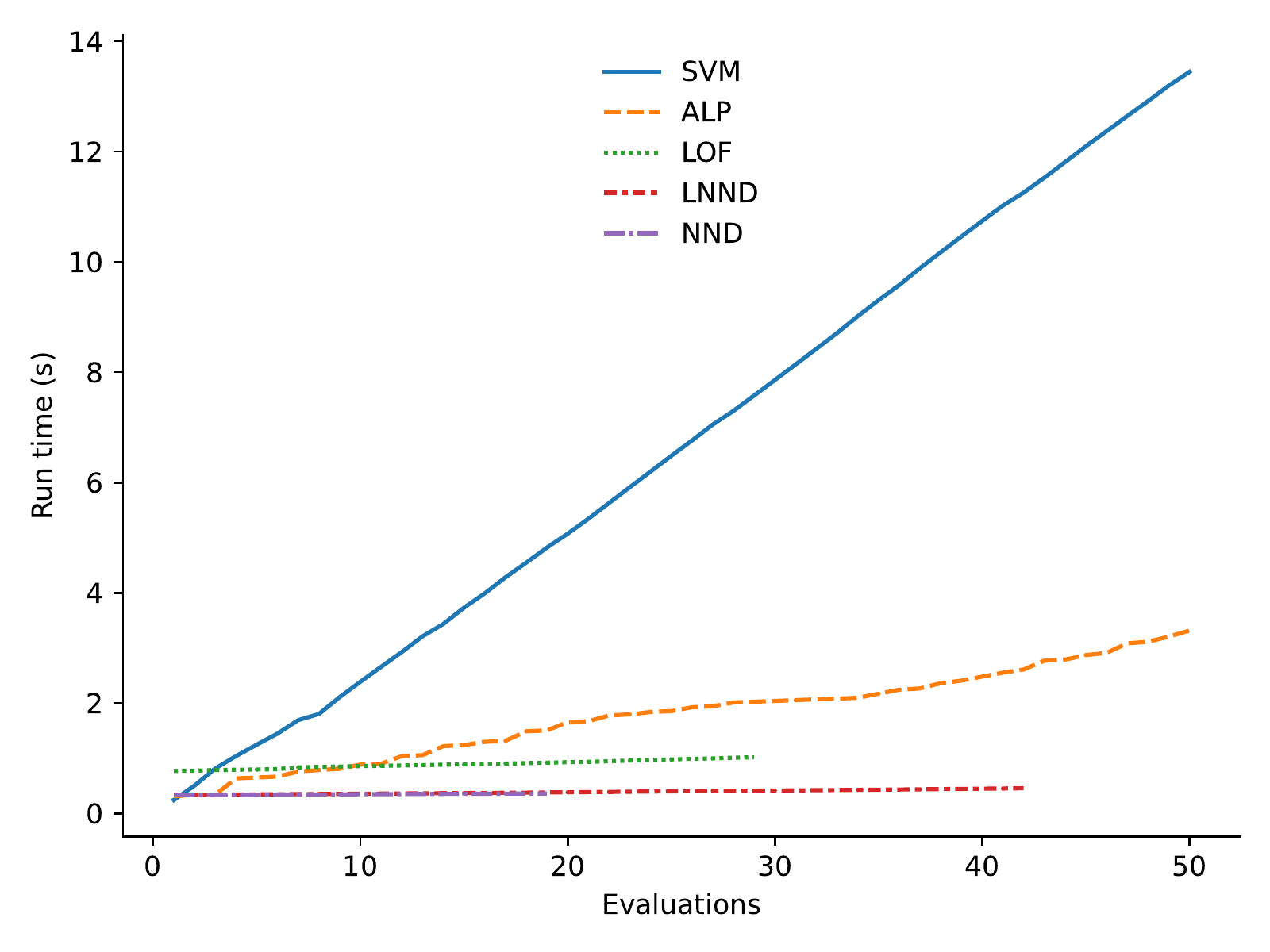}%
\caption{Mean run times (5 runs) of hyperparameter optimisation with Malherbe-Powell on a training set with 1000 target class instances and 1000 other instances, drawn from the \e{miniboone} dataset (target class 1).}
\label{fig_optimisation_run_times}
\end{figure}

Finally, Figure \ref{fig_optimisation_run_times} displays the run times of hyperparameter optimisation as a function of the number of evaluations. These run times are implementation-dependent, but we can nevertheless make a number of broad observations. For SVM, run time is directly proportional to the number of evaluations, as no calculations are reused. NND, LNND and LOF can effectively be optimised in constant time, since the initial nearest neighbour queries dominate. The run time of LOF is higher because it uses five-fold cross validation and needs five nearest neighbour queries. For ALP, we observe a considerable amount of additional run time per evaluation. Looking at individual evaluations, we find that their run time varies wildly, seemingly due to the computational load of working with large arrays when $k$ and $l$ are large.

The higher run time required by SVM for additional evaluations is compounded by the finding, reported above, that optimisation of SVM requires more evaluations than ALP, LOF, LNND and especially NND. This is illustrated by the fact that the curves of the last three data descriptors in Figure \ref{fig_optimisation_run_times} end before 50 evaluations. 

\section{Conclusion}
\label{sec_conclusion}
In this paper, we have presented a thorough analysis of hyperparameter optimisation for five data descriptors, SVM, NND, LNND, LOF and ALP. We have explained how NND, LNND and ALP can be optimised efficiently with a single nearest neighbour query and leave-one-out validation, while SVM requires building a new model for each additional hyperparameter evaluation. We then evaluated the performance of hyperparameter optimisation empirically, based on a large selection of 246 one-class classification problems drawn from 50 datasets. 

From a selection of optimisation algorithms, the recent Malherbe-Powell approach provides the best overall performance with all five data descriptors. LNND is most sensitive to overfitting, followed by LOF, ALP, SVM and NND, but in all cases overfitting reduces with target set size. For all data descriptors, optimised hyperparameters significantly outperform default hyperparameter values after a handful of evaluations. As predicted, different hyperparameter values can be evaluated more efficiently for NND, LNND, LOF and ALP than for SVM. In addition, these data descriptors also require fewer evaluations than SVM. After 50 evaluations, ALP and SVM significantly outperform LOF, NND and LNND, and LOF and NND in turn perform better than LNND. SVM also outperforms ALP on our datasets, but the difference is not significant.

A more detailed look at the difference between ALP and SVM revealed that their strengths are to some extent complementary, and that selecting one or the other based on their validation AUROC gives the best results. SVM has a strong relative advantage with difficult one-class classification problems, while ALP performs better with problems with which a good AUROC of 0.8 or higher can be achieved. 

Overall, we come to the following conclusion. NND is a very simple data descriptor that can be optimised very efficiently. While the resulting gain in performance is limited, it nevertheless leads to results that are generally better than what can be obtained with a data descriptor with default hyperparameter values. SVM is a data descriptor with excellent performance, but it is expensive to optimise. The performance of ALP rivals that of SVM, and potentially surpasses it with one-class classification problems that admit a good solution, but it can be optimised much more efficiently.

Therefore, we find that ALP is a good default choice, while NND may appeal to practitioners constrained by a smaller computational budget. If the absolute best performance is desired, we recommend that practitioners consider both ALP and SVM, and make the choice dependent on validation AUROC.

In future research, we think that it could be worthwhile to investigate in greater detail what properties of datasets determine the relative strengths and weaknesses of ALP and SVM. A deeper understanding of this question could in turn be applied to modify the ALP and SVM algorithms. In addition, it would be useful if the computational cost of optimising SVM and ALP could be reduced.

We have focused our attention in this paper on a handful of hyperparameters with the most immediate impact on the classification of different datasets. But these are not the only choices available to a practitioner. Hyperparameter optimisation is a specific form of model selection, and conversely, any modification to a classification algorithm can be seen as a hyperparameter choice. In particular, one large topic that we have set aside in the present paper is the possibility to change how similarity and difference are measured, by choosing a different metric, kernel and/or scaling function. These are essentially open-ended choices, so part of the challenge lies in delineating the search area. 

In the nearby future, we plan to investigate how the optimisation of hyperparameters can best be integrated into data descriptor ensembles in a multi-class setting.

\section*{Acknowledgements}
\label{sec_acknowledgement}
The research reported in this paper was conducted with the financial support of the Odysseus programme of the Research Foundation -- Flanders (FWO). D. Peralta is a Postdoctoral Fellow of the Research Foundation -- Flanders (FWO, 170303/12X1619N).

\section*{Conflict of interest}
The authors declare that they have no conflict of interest.

\bibliographystyle{spbasic}
\bibliography{20201103_oc_tuning}

\end{document}